\pdfoutput=1

\documentclass[11pt]{article}

\usepackage[]{EMNLP2023}

\usepackage{times}
\usepackage{latexsym}

\usepackage{algorithm}
\usepackage{multirow}
\usepackage{graphicx}
\usepackage{subfigure}
\usepackage{amsmath}
\usepackage{enumitem}
\usepackage{booktabs}
\usepackage{xspace}
\usepackage{microtype}
\usepackage{multicol}
\usepackage{caption}
\usepackage{amssymb}
\usepackage{array}
\usepackage{fp}
\usepackage{makecell}
\usepackage{flushend}
\usepackage{cases}
\usepackage{color}
\usepackage{algpseudocode}
\usepackage{listings}
\usepackage{xcolor}
\usepackage{mathrsfs}
\usepackage{pifont}

\newcommand{\paratitle}[1]{\vspace{1.5ex}\noindent\textbf{#1}}
\newcommand{\ie}{\emph{i.e.,}\xspace}

\newcommand{\eg}{\emph{e.g.,}\xspace}

\newcommand{\ignore}[1]{}

\newcommand{\cmark}{\ding{51}}
\newcommand{\xmark}{\ding{55}}

\usepackage[T1]{fontenc}

\usepackage[utf8]{inputenc}

\usepackage{microtype}

\usepackage{inconsolata}

\title{KG-Agent: An Efficient Autonomous Agent Framework for Complex Reasoning over Knowledge Graph}

\author{
    \textbf{Jinhao Jiang\textsuperscript{{1},{3}},
            Kun Zhou\textsuperscript{{2},{3}},
            Wayne Xin Zhao\textsuperscript{{1},{3}}\thanks{\llap{}\:\:\:Corresponding author.}, 
            Yang Song\textsuperscript{{4}}\footnotemark[1],
            }\\ 
    \textbf{Chen Zhu\textsuperscript{{5}}, Hengshu Zhu\textsuperscript{{5}}, Ji-Rong Wen\textsuperscript{{1},{2},{3}}}\\
	\textsuperscript{1}Gaoling School of Artificial Intelligence, Renmin University of China.\\
	\textsuperscript{2}School of Information, Renmin University of China.\\
	\textsuperscript{3}Beijing Key Laboratory of Big Data Management and Analysis Methods.\\
	\textsuperscript{4}NLP Center, BOSS Zhipin.
        \textsuperscript{5}Career Science Lab, BOSS Zhipin.\\
    	\texttt{jiangjinhao@ruc.edu.cn, batmanfly@gmail.com}\\
}

\begin{document}
\maketitle

\begin{abstract}
In this paper, we aim to improve the reasoning ability of large language models~(LLMs) over knowledge graphs~(KGs) to answer complex questions.
Inspired by existing methods that design the interaction strategy between LLMs and KG, we propose an autonomous LLM-based agent framework, called \textbf{KG-Agent}, which enables a small LLM to actively make decisions until finishing the reasoning process over KGs.
In KG-Agent, we integrate the LLM, multifunctional toolbox, KG-based executor, and knowledge memory, and develop an iteration mechanism that autonomously selects the tool then updates the memory for reasoning over KG.
To guarantee the effectiveness, we leverage program language to formulate the multi-hop reasoning process over the KG, and synthesize a code-based instruction dataset to fine-tune the base LLM.
Extensive experiments demonstrate that only using 10K samples for tuning LLaMA-7B can outperform state-of-the-art methods using larger LLMs or more data, on both in-domain and out-domain datasets. 
Our code and data will be publicly released.
\end{abstract}

\section{Introduction}
\label{introduction}
Despite the remarkable performance on various NLP tasks~\cite{brown-NIPS-20,zhao-arxiv-23}, large language models~(LLMs) still have limited capacities in solving complex tasks~\cite{Hu-arxiv-2023-do} solely based on their parametric knowledge, \eg multi-hop and knowledge-intensive reasoning~\cite{Lan-KBQA_survey-2023}.
Knowledge graph~(KG), which stores massive knowledge triples in a graph-structured format, has been broadly used to complement LLMs with external knowledge~\cite{Pan-arxiv-unifying}.

Due to the large volume and structured format of KG data, it is not easy for LLMs to effectively utilize the information from KG. 
Recent work mainly adopts \textit{retrieval-augmented}~\cite{RNG-KBQA} or \textit{synergy-augmented}~\cite{Jiang-2023-arxiv-StructGPT} methods to enhance LLMs with KG data.
The former approach retrieves and serializes the task-related triples as part of the prompt for LLMs, while the latter approach designs an information interaction mechanism between KG and LLMs to iteratively find the solution to the question. 
In particular, synergy-augmented methods can benefit from the structured search on KG (\eg SPARQL) and the language understanding capacity of LLMs, achieving comparable or even better performance compared with previous state-of-the-art methods~\cite{Gu-ACL-23}. 

However, there are still two major limitations on existing synergy-augmented methods. 
First, the information interaction mechanism between LLM and KG is often pre-defined (\eg following a human-crafted multi-round plan), which cannot flexibly adapt to various complex tasks~\cite{Luo-arxiv-23-Reasoning, Jiang-2023-arxiv-StructGPT}. For instance, it would become ineffective to handle the unintended requirements in the reasoning process, \eg varied difficulties or constraints.  
Second, these methods~\cite{wang-arxiv-2023-a} mostly rely on stronger closed-source LLM APIs (\eg ChatGPT and GPT-4) to understand or learn to solve complex tasks. However, the distilled plans or procedures, also limited to special task settings or capacity levels,  may not be best suited for instructing these weaker models. 

To address these issues, in this paper, we propose the \textbf{KG-Agent}, an autonomous LLM-based agent framework for complex reasoning tasks over KG. 
The motivations are twofold: (1) designing autonomous reasoning approaches that can actively make decisions during reasoning, without human assistance; (2) enabling relatively small models (\eg 7B LLM) to effectively perform complex reasoning,  without reliance on close-sourced LLM APIs. 
To achieve this, our approach makes three major technical contributions. First, we extend the LLM's capacity to manipulate structured data by curating a multifunctional toolbox, enabling LLM to perform discrete or advanced operations (\eg filtering, counting, and retrieval) on KG data and intermediate results. Second, we leverage existing KG reasoning datasets for synthesizing code-based instruction data to fine-tune the LLM, where we first generate the program according to the reasoning chain on KG and then synthesize the instruction data. 
Third, we propose an autonomous iteration mechanism based on tool selection and memory updation that integrates the tuned LLM, multifunctional toolbox, KG-based executor, and knowledge memory, for autonomously reasoning over KG.
 
To verify the effectiveness, we evaluate  KG-Agent on both in-domain and out-of-domain tasks including KG-based question answering~(KGQA) and open domain question answering~(ODQA).
With much fewer training data (\ie 10K samples) for tuning a smaller LLM (\ie LLaMA-7B), our approach can outperform competitive LLM-based baselines on in-domain datasets~(\eg using about 36\% and 23\% of the original training set amount while obtaining 7.5\% and 2.7\% relative improvement of F1 on CWQ and GrailQA respectively).
On the out-of-domain datasets, the zero-shot performance of our KG-Agent is better than competitive full-data supervised fine-tuning models~(\eg 9.7\% and 8.5\% relative improvement of accuracy on WQ-Freebase and TQ-Wiki, respectively).

\section{Related Work}
\label{related_work}
\paragraph{LLM-based KG Reasoning.}
Benefitting from the powerful zero-shot and few-shot capability, recent studies have leveraged LLMs to perform reasoning over KG.
Recent work can be roughly divided into \textit{retrieval-augmented}~\cite{TIARA} and \textit{synergy-augmented}~\cite{Gu-ACL-23} two types. The retrieval-augmented method is to retrieve and serialize the triples from the KG, and then feed it to the LLM to help generate the final results~(\eg answers or SPARQL query)~\cite{RNG-KBQA}. Such a way loses the structured information in the original KG and may retrieve redundant knowledge, limiting LLMs' understanding. To relieve these problems, the synergy-augmented methods design an information interaction mechanism between LLMs and KGs to enable LLMs to query KGs multiple times to answer the question~\cite{Jiang-2023-arxiv-StructGPT}. Specifically, they either first generate the full plan~\cite{Li-ICLKBQA-arXiv-23} and then ground it on KG, or make a plan step-by-step based on the KG~\cite{Luo-arxiv-23-Reasoning}. Although obtaining better performance, the information interaction mechanism in existing methods often follows a pre-defined way, which cannot flexibly adapt to various complex tasks. In contrast, our proposed KG-Agent can autonomously make decisions during reasoning over KG, without human assistance.

\paragraph{LLM-based Agents.}
Recently, LLMs have shown surprising long-horizon planning and reasoning capabilities~\cite{Reflexion, MemoryBank}, and LLM-based agents have gradually become a hot topic for autonomously solving complex interactive tasks~\cite{agent-survey}.
A large number of agents focus on general-purpose task solving. As the representative projects, ReAct~\cite{ReAct} proposes a prompting method to convert LLMs~(\eg ChatGPT) as language agents, to interact with the external environment, receive the feedback, and then generate the action for next step reasoning. Then, AutoGPT\footnote{https://github.com/Significant-Gravitas/AutoGPT} further empowers LLMs~(\ie GPT4) with long/short-term memory management and external tools like search engines to autonomously address a user request.
In addition, several other agents also focus on specific domains, such as WebGPT~\cite{WebGPT} for the web-browsing environment, MM-REACT~\cite{MM-REACT} for the multi-modal scenario, and ProgPrompt~\cite{ProgPrompt} for the real-life environment.
However, recent works involving language agents mostly rely on stronger closed-source LLM APIs (\eg ChatGPT and GPT-4) to understand or learn to solve complex tasks.
Our KG-Agent is the first autonomous agent framework to support complex reasoning over KG only relying on a relatively smaller 7B LLM.

\section{Preliminary}
\label{preliminary}

In this section, we first formally define knowledge graph~(KG), and then formalize the complex knowledge reasoning task based on KG.

\paratitle{Knowledge Graph~(KG).} 
A knowledge graph typically consists of a large number of fact triples, expressed as $\mathcal{G} = \{\langle e,r,e' \rangle| e,e' \in \mathcal{E}, r \in \mathcal{R}\}$, where $\mathcal{E}$ and $\mathcal{R}$ denote the entity set and relation set, respectively.
A triple $\langle e,r,e' \rangle$ describes a factual knowledge that a relation $r$ exists between the head entity $e$ and tail entity $e'$. 
Each entity $e$ is assigned a unique entity ID (or string value), and belongs to one entity type $t$ such as \emph{Country} and \emph{Person}.  
Furthermore, we introduce \emph{neighboring relations} to denote both the incoming and outgoing relations for a set of entities $\{e\}$, denoted as $\mathcal{R}_{\{e\}} = \{r | \langle e, r, e' \rangle \in \mathcal{G}\} \cup \{r | \langle e', r, e \rangle \in \mathcal{G} \}$.

\paratitle{Problem Formulation.} 
In this work, we assume that a KG is available and contains the answer entities for the given natural language question.
Our objective is to develop a LLM-based agent that can autonomously infer the answer to the question based on the given KG. 
As it has been shown that domain-specific interface is helpful for LLMs to manipulate the structured data~\cite{Jiang-2023-arxiv-StructGPT}, we further assume that a toolbox can be provided to facilitate the access to the information of KG. 
Formally, given a natural language question $q$, and a toolbox $\mathcal{T}$ and a KG $\mathcal{G}$, we aim to develop a capable agent to deduce the final answers $A_q=\{e\}$ for the question $q$ by leveraging the tools in $\mathcal{T}$ and the knowledge information in $\mathcal{G}$.

\section{Approach}
\label{approach}

\begin{figure*}[t]
    \centering
    \includegraphics[width=2.1\columnwidth]{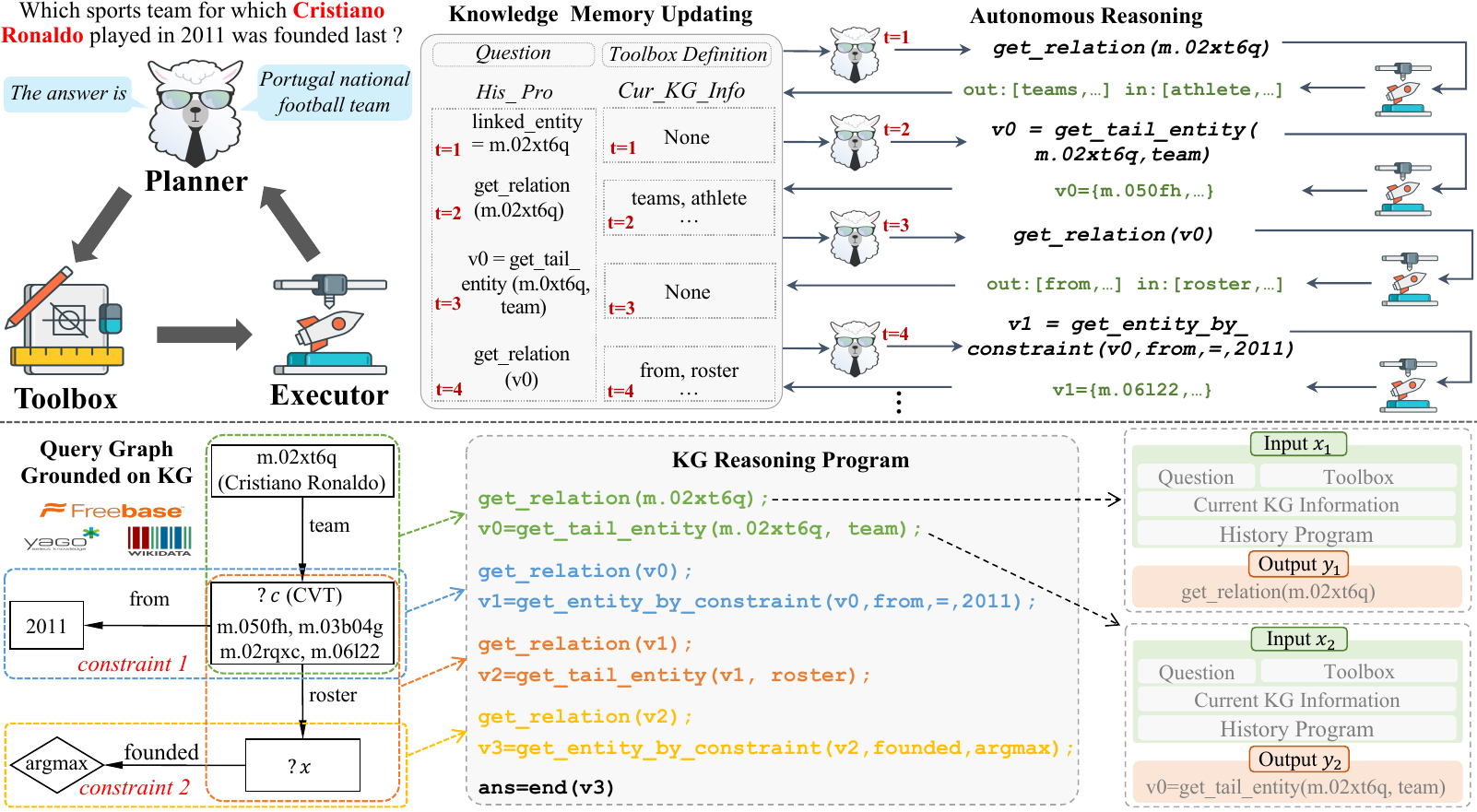}
    \caption{The overview of our proposed KG-Agent. The top half is the workflow of our agent, and the bottom half is an example of instruction fine-tuning data synthesis and the prompt template for the input-output pairs. For brevity, we simplify the relation surface form.}
    \label{fig:model}
\end{figure*}

In this part, we present the proposed \textbf{KG-Agent} for autonomously solving complex reasoning tasks over KG. The core of our KG-Agent framework is a well-instructed LLM, which can autonomously make decisions when reasoning over KG.
We first extend the LLM's capacities by designing a toolbox with supporting tools to manipulate the KG data or intermediate results (Section~\ref{sec:toolbox}). 
To enhance the step-by-step reasoning capacity, we leverage existing knowlege graph question answering~(KGQA) datasets to synthesize KG reasoning programs and convert them into formatted instruction tuning data (Section~\ref{sec:tuning}). Finally, we design an effective agent framework based on the knowledge memory to support autonomous reasoning over KG (Section~\ref{sec:utilization}).  
Next, we give the technical details of KG-Agent.  

\subsection{Toolbox for Knowledge Graph}
\label{sec:toolbox}
Since LLMs struggle to accurately manipulate the structured data~\cite{Jiang-2023-arxiv-StructGPT}, we construct a supporting toolbox for easing the utilization of the KG information.
According to existing work~\cite{GrailQA,KQA-Pro}, reasoning over KG ~(\eg Freebase or Wikidata) typically requires three fundamental operations, namely extracting information from KG, filtering irrelevant information based on the semantics of the question, and operating on the extracted information.
Therefore, we design three types of tools for LLMs reasoning over KG, \ie extraction, semantic, and logic tools.

$\bullet$ \textbf{Extraction tools} aim to facilitate the access to information from KG. 
Considering the basic data types in  KG, we design five tools to support the access to the relations (\emph{get\_relation}), the head/tail entities (\emph{get\_head\_entity}/\emph{get\_tail\_entity}), and entities with specific type or constraint (\emph{get\_entity\_by\_type}/ \emph{get\_entity\_by\_constraint}), \emph{w.r.t.} some entity set or other input information (\eg relation or type).  

$\bullet$ \textbf{Logic tools} aim to 
support basic manipulation operations on the extracted KG information, including entity counting (\emph{count}), entity set intersection (\emph{intersect}) and union (\emph{union}), condition verification (\emph{judge}), and ending the reasoning process with the current entity set as the final answer(s) (\emph{end}). 

$\bullet$ \textbf{Semantic tools}
are developed by utilizing pre-trained models to implement specific functions, including relation retrieval (\emph{retrieve\_relation}) and entity disambiguation (\emph{disambiguate\_entity}). 
These tools extend the basic operations on KGs and can support advanced functionalities for KG reasoning. 

We summarize the detailed definition and usage of the tools in Table~\ref{tab:toolbox} at the Appendix~\ref{sec:app-tool}. Note that since the format and usage for each tool have been defined in a unified way, the toolbox for KG can be flexibly extended according to the real needs.

\subsection{KG-Agent Instruction  Tuning}
\label{sec:tuning}
To enable the autonomous reasoning process, we construct a high-quality instruction dataset for fine-tuning a small LLM (\ie LLaMA2-7B).
For this purpose, we first leverage existing KG based question answering~(KGQA) datasets to generate the KG reasoning program, and then decompose it into multiple steps. Finally, each step is formulated as the instruction data with input and output.

\subsubsection{KG Reasoning Program Generation}\label{sec-KG-RPG}
Instead of distilling from close-sourced LLMs (\eg GPT-4),  we propose to leverage existing KGQA datasets to synthesize the KG reasoning program. 
These KGQA datasets contain the annotated SQL queries that can be executed to directly extract the answer entities for each question. 
In particular, the SQL query generally includes the relation chain, conditions, or constraints, which are beneficial for reasoning program synthesis.  
Concretely, we first ground the SQL query on the KG to obtain a query graph, then extract the reasoning chain and constraint conditions from the query graph, and finally decompose the chain into multiple code snippets as the reasoning program.

\paratitle{Reasoning Chain Extraction.}
Since the whole KG is extremely large and contains irrelevant data, the first step is to acquire a small KG subgraph related to the question, referred to as \emph{query graph}. Following previous work~\cite{Yin-acl-20}, we obtain the query graph from the KG via rule match.
As shown in Figure~\ref{fig:model}~(b), the query graph has a tree-like structure that can be directly mapped to a logical form~\cite{Yin-acl-20}, and it can clearly depict the execution flow of the SQL query to obtain the answer. 
Second, starting from the mentioned entity in the question (\ie \emph{Cristiano Ronaldo}), we adopt breadth-first search~(BFS) to visit all the nodes on the query graph. 
This strategy would finally produce a reasoning chain~(\eg \textit{teams$\rightarrow$roster\_team}) linking the start entity to the answer entity, and the relevant constraint conditions~(\eg \textit{roster\_from} $=$ ``2011'') or numerical operation~(\eg \textit{founded} must be last) can be naturally involved in this process. 

\paratitle{Reasoning Program Generation.} After extracting the reasoning chain, we next convert it into multiple interrelated triples, where each triple generally corresponds to an intermediate reasoning step.
Finally, we reformulate the triples into several function calls with the code format, which represents the tool invocation and can be executed to obtain the corresponding triples based on the KG.
Given a triple $\langle e, r, e'\rangle$, we craft a rule-based method to synthesize the function calls that represent the information flow from $e$ to $e'$.
Specifically, we start from the \emph{get\_relation(e)} function call to obtain the current candidate relations $\{r\}$ associated with $e$ on the KG. Then, we select one relation $r$ and pass it to other required function calls~(\eg \textit{get\_tail\_entity} or \textit{get\_entity\_by\_constraint}), and finally obtain new entities.
Following the order of the reasoning chain, we generate all the function calls to compose the final KG reasoning program for producing the instruction dataset.
We show one example in Figure~\ref{fig:model}~(b) to intuitively illustrate the conversion process from the annotated SQL query to our required KG reasoning program.

\subsubsection{KG Reasoning Instruction Synthesis}\label{sec-instruction-syn}
After obtaining the reasoning program on KG, we further utilize it for synthesizing instruction data for supervised fine-tuning (SFT).  
As discussed in Section~\ref{sec-KG-RPG}, our instruction data is highly based on the reasoning program, which is aligned with the intermediate reasoning steps for KGQA.  

\paratitle{Input-Output Pair Construction.}
The synthetic KG reasoning program consists of multiple function calls in a sequence. For each function call, we aim to construct an input-output pair as the instruction. Specifically, the input contains the question, toolbox definition, current KG information (\ie the next candidate relations of the current entity set), and history reasoning program before the current step; and the output is the function call at the current step.
Next, after executing the function call at the current reasoning step, the history reasoning program and current KG information in the input will be accordingly updated, and the output will be updated as the function call at the next step.
By iterating the above process, for each sample in the KGQA datasets, we can obtain multiple input-output pairs derived from the corresponding reasoning program, which depict the complete reasoning trajectory on the KG.
To help LLMs better understand, we further utilize a unified prompt, as shown in Figure~\ref{fig:model}~(c), to format each input-output pair and obtain the final instruction tuning data.

\paratitle{Agent Instruction Tuning.}
Based on the above formatted instruction tuning data, we perform supervised fine-tuning on a small LLM~(\ie LLaMA-7B), which is much smaller than the backbone models in previous work~\cite{Jiang-2023-arxiv-StructGPT}.
Formally, for each sample, we formulate all input-output pairs of the complete trajectory in the format of $\{\langle x_1, y_1 \rangle , ..., \langle x_t, y_t \rangle, ..., \langle x_{n}, y_{n} \rangle \}$, where $\langle x_t, y_t \rangle$ represent the input and ground-truth response in the $t$-th step and $n$ represents the total steps.
For simplicity, we denote each input and output as $x$ and $y$ below.
During the instruction tuning process, we feed the input~$x$ and output~$y$ into the decoder-only LLM and minimize the cross-entropy loss on the ground-truth response~$y$ as:
\begin{align}
    \mathcal{L} = - \sum_{k=1}^m \log \text{Pr}(y_k|x, y_{<k}),
\end{align}
where $m$ denotes the number of tokens in $y$, $y_k$ and $y_{<k}$ are the $k$-th and previous tokens in the output.

\begin{table}[t]
        \renewcommand\arraystretch{1.0}
        \setlength\tabcolsep{2.5pt}
	\centering
	\small
            \begin{tabular}{l |  c | c c c c }
			\toprule
			\textbf{Method} & \textbf{\makecell{Work\\ Flow}} & \textbf{\makecell{Base \\Model}} & \textbf{Tool} & \textbf{Memory} & \textbf{\makecell{Multi\\Task}} \\
			\midrule
			Pangu & pd & T5-3B & \xmark & \xmark & \xmark\\
                StructGPT & pd & ChatGPT  & \cmark  & \xmark & \xmark\\
			RoG & pd & LLaMA-7B & \xmark & \xmark & \xmark\\
                ChatDB & auto & ChatGPT & \xmark & \cmark & \xmark\\
                KB-BINDER & pd & CodeX & \xmark & \xmark & \xmark\\
                \midrule
                KG-Agent & auto & LLaMA2-7B & \cmark & \cmark & \cmark\\
    		\bottomrule
		\end{tabular}
        \caption{Comparison of different methods. \textbf{\emph{Work Flow}} describes that the interaction way between the LLM and KG is pre-defined~(``pd'') or autonomous~(``auto''). \textbf{\emph{Multi Task}} means whether to support generalization across different KGs via multi-task learning.}
	\label{tab:res}
    \vspace{-0.3cm}
\end{table}

\subsection{Autonomous Reasoning over KG}
\label{sec:utilization}
After instruction tuning, we further design an effective agent framework
that enables KG-Agent to autonomously perform multi-step reasoning over KG for answer finding.
The overall illustration of KG-Agent is shown in Figure~\ref{fig:model}~(a).
It mainly contains four components, \ie the core instruction-tuned LLM (Section~\ref{sec:tuning}), referred to as the \emph{LLM-based planner}, the multifunctional \emph{toolbox}~(Section~\ref{sec:toolbox}), the \emph{KG-based executor} for executing the tool invocation, and the \emph{knowledge memory} to record the context and currently useful information in the whole process.
Next, we introduce how KG-Agent performs autonomous reasoning over KG.

\paratitle{Knolwedge Memory Initialization.}
The knowledge memory preserves the currently useful information to support the LLM-based planner for making decisions.
It mainly contains four parts of information, \ie natural language question, toolbox definition, current KG information, and history reasoning program.
The former two parts are initialized with the given question and toolbox definition, which remain unchanged during the reasoning process.
The later two parts are initialized as an empty list, which will be constantly updated at each step after LLM generating the function call and executor invoking the corresponding tool.

\paratitle{Planner for Tool Selection.}
Based on the current knowledge memory, the LLM-based planner selects a tool to interact with KG at each step.
Specifically, all the parts in the current knowledge memory will be formatted with corresponding prompt template to compose the input (used in Agent Instruction Tuning in Section~\ref{sec-instruction-syn}), and then the LLM will generate one function call by selecting a tool and its arguments from the input. 
Generally, the planner needs to invoke tools from the pre-defined toolbox to address four types of task requirements, \ie linking the mentioned entity to KG (\eg ``\emph{get\_candidate\_entity}'' and ``\emph{disambiguate\_entity}''), accessing the KG information (\eg \textit{``get\_relation''} and \textit{``get\_head\_entity''}), processing the intermediate results (\eg \textit{``count''} and \textit{``intersect''}), or returning the final answer to end the reasoning process (\eg \textit{``end''}).

\begin{table*}
\small
\centering
\resizebox{\textwidth}{!}{\begin{tabular*}{\textwidth}{l@{\extracolsep{\fill}}cccccccc}
\toprule
\multirow{2.5}{*}{\textbf{Model}}& \multicolumn{2}{c}{\textbf{WebQSP}}   &\multicolumn{2}{c}{\textbf{CWQ}} &\multicolumn{4}{c}{\textbf{GrailQA~(F1)}} \\ 
\cmidrule{2-3}\cmidrule{4-5}\cmidrule{6-9}
& \textbf{Hits@1} & \textbf{F1} & \textbf{Hits@1} & \textbf{F1} & \textbf{Overall} &  \textbf{I.I.D.} & \textbf{Compositional} & \textbf{Zero-shot}  \\
\midrule
GraftNet     & 66.4  &  60.4     & 36.8  & 32.7      & - &   -    & -  &  -   \\
NSM      & 68.7  &  62.8     & 47.6  & 42.4    & -   & -    & -  &  -   \\
SubgraphRetrieval   & 69.5& 64.1  & 49.3& 46.3 &  - & - & -  &  -   \\
UniKGQA      &  75.1& 70.2  & 50.7 & 48.0 &  - & -    & -  &  -  \\
ReasoningLM &  78.5& 71.0  & 69.0 & 64.9 &  - & -    & -  &  -  \\
\midrule
RNG-KBQA    & -  &  75.6     & -  & -    & 76.8  &   89.0    & 68.9  &  74.7   \\
Uni-Parser & -  &  75.8     & -  & -    & 76.5  &   88.3    & 71.4  &  73.4   \\
ArcaneQA & - &75.6 & -  & -       & 76.9  &   89.2    & 73.9  &  72.8 \\
PanGu w/ T5-3B & - & 79.6 & - & - & 83.4 & - & - & - \\
TIARA & 75.2 & 78.9 & -  & -       &  81.9  &   91.2    & 74.8  &  80.7 \\
FC-KBQA & - & 76.9 & - & 56.4 & 83.8 & 91.5 & 77.3 &  83.1 \\
\midrule
ROG & \textbf{85.7} & 70.8 & 62.6 & 56.2 & -  & - & - & - \\
ChatGPT & 67.4 & 59.3 & 47.5 & 43.2 & 25.3 & 19.6 & 17.0 & 31.2 \\
Davinci-003 & 70.8 & 63.9 & 51.4 & 47.6 & 30.1 & 23.5 & 22.0 & 36.4 \\
GPT-4 & 73.2 & 62.3 & 55.6 & 49.9 & 31.7 & 25.0 & 20.6 & 39.2 \\
StructGPT & 72.6 & 63.7 & 54.3 & 49.6 & 54.6 & 70.4 & 44.3 & 50.5 \\
\midrule
\multicolumn{1}{l}{Ours} & 83.3 & \textbf{81.0}  & \textbf{72.2} & \textbf{69.8} & \textbf{86.1} & \textbf{92.0} & \textbf{80.0} & \textbf{86.3} \\
\bottomrule
\end{tabular*}}
\caption{The results on the test set of WebQSP and CWQ, and dev set of GrailQA, which are based on Freebase KG. We copy part of the results from~\citet{Jiang-2023-arxiv-StructGPT,Gu-ACL-23,Luo-arxiv-23-Reasoning} and evaluate ChatGPT,Davinci-003, GPT-4, and StructGPT with OpenAI API. Bold font denotes the best performance.}
\label{tab:fb-res}
\end{table*}

\paratitle{Executor for Memory Updation.}
After the planner generates the function call, the KG-based executor will execute it using a program compiler. It can cache or operate the intermediate variables, and extract new entities or relations from the KG. After execution, the knowledge memory will be accordingly updated. First, the current function call will be added to the history reasoning program. Second, if the invoked tool is to obtain the new information from the KG (\eg ``\emph{get\_relation}''), 
the executor will add it to the KG information for updating the knowledge memory.

\paratitle{Iterative Autonomous KG-Agent.}
The KG-Agent framework autonomously iterates the above tool selection and memory updation process to perform step-by-step reasoning, where the knowledge memory is used to maintain the accessed information from KG. 
In this way, the multi-turn decision-making process of the agent is like walking on the KG along relations.
Once reaching the answer entities, the agent will automatically stop the iterative process.
Note that the whole process is agnostic to the task types (\eg question answering) and some specific KGs. Therefore, our approach is a general framework that can be applied to a variety of complex tasks that require reasoning over any KGs.

\subsection{Comparison to Previous Work}

Existing methods of reasoning over KG can be categorized into two classes based on their workflow. The first line of research, such as KB-BINDER~\cite{Li-ICLKBQA-arXiv-23}, Pangu~\cite{Gu-ACL-23}, StructGPT~\cite{Jiang-2023-arxiv-StructGPT}, and RoG~\cite{Luo-arxiv-23-Reasoning}, crafted a pre-defined interaction way between LLM and KG, which cannot flexibly adapt to various complex tasks. Another line of research, such as ChatBD~\cite{ChatDB}, conducted autonomous reasoning with chain-of-thought and memory augmented. However, it relies on the strong closed-source LLM APIs~(\eg ChatGPT) and cannot use tools to implement some specialized operations~(\eg count). Our KG-Agent is the first autonomous agent framework to support the complex interaction between LLM and KG with tool and memory augmented. Furthermore, we implement this autonomous agent by instruction tuning a smaller 7B open-source LLM compared to the backbone LLM in KB-BINDER, StructGPT, and ChatDB.
At the same time, the agent instruction tuning data is constructed from various KGs~(\eg Wikidata and Freebase), which helps our KG-Agent to learn the general autonomous decision making capabilities over various KGs.

\begin{table*}
\small
\centering
\resizebox{\textwidth}{!}{\begin{tabular*}{\textwidth}{l@{\extracolsep{\fill}}cccccccc}
\toprule
\textbf{Model}& \textbf{Overall} & \textbf{Multi-hop}   & \textbf{Qualifier} & \textbf{Comparison} & \textbf{Logical}    & \textbf{Count}  & \textbf{Verify} & \textbf{Zero-shot}   \\ 
\midrule
KVMemNet &16.61  &16.50 &18.47 &1.17 &14.99& 27.31 &54.70& 0.06   \\
EmbedKGQA &28.36 &26.41 &25.20 &11.93 &23.95 &32.88 &61.05 &0.06 \\
RGCN &35.07 &34.00 &27.61 &30.03 &35.85 &41.91 &65.88 &0.00 \\
\midrule
RNN SPARQL &41.98& 36.01 &19.04& 66.98& 37.74 &50.26 &58.84& 26.08   \\
BART SPARQL &89.68 & 88.49 &83.09 &96.12 &88.67 &85.78 &92.33& 87.88 \\
\midrule
ChatGPT & 24.96 & 24.22 & 26.37 & 39.15 & 25.51 & 10.76 & 54.70 & 15.67 \\
Davinci-003 & 31.02 & 29.58 & 31.58 & 49.8& 29.62 & 16.70 & 65.54 & 21.83 \\
GPT-4 & 37.43 & 34.82 & 37.15 & 55.75 & 36.81 & 15.27 & 72.93 & 27.28  \\
\midrule 
Ours & \textbf{92.15} & \textbf{91.03} & \textbf{87.90} & \textbf{96.32} & \textbf{91.28 }& \textbf{88.21} & \textbf{92.86} & \textbf{91.40} \\
\bottomrule
\end{tabular*}}
\caption{The accuracy on the test set of KQA Pro, which is based on Wikidata KG. The results of Davinci-002, GPT-4, and ChatGPT are evaluated by us and the results of other baselines are copied from~\citet{KQA-Pro}.}
\label{tab:wiki-res}
\end{table*}

\section{Experiment}
\label{experiment}
\subsection{Experimental Setup}
\label{sec:app_experiment}
We select four commonly-used KGQA datasets as in-domain datasets, \ie \textit{WebQSP}, \textit{CWQ}, and \textit{GrailQA}, which are based on Freebase, and \textit{KQA Pro}, which is based on Wikidata. And we select three ODQA datasets as out-of-domain datasets, \ie \textit{WQ}, \textit{NQ}, and \textit{TQ}.
Further, we consider three types of baseline methods, \ie \textit{subgraph-based reasoning}, \textit{LM-based seq2seq generation}, and \textit{LLM-based methods} for comparison on in-domain datasets, and \textit{Fine-tune based} and \textit{LLM-based} methods for out-of-domain datasets.
We show the details of the above datasets, baselines, evaluation protocol, and implementation in Appendix~\ref{sec:app-experiment}.

\begin{table}[t]
	\centering
	\small
            \begin{tabular}{l  c  c c }
			\toprule
			\textbf{Models} & \textbf{NQ-Wiki} & \textbf{TQ-Wiki} & \textbf{WQ-Freebase} \\
			\midrule
			T5-Base & 30.94 & 27.63 & 24.06 \\
			T5-Large & 31.21 & 29.40 &  {24.70} \\
			BART-Base & 29.47 & 25.43 & 21.95 \\
			BART-Large	& 32.60 & 33.05 & 26.33 \\
                \midrule
                Davinci-003  & 51.94 & 88.57 & 23.81 \\
			ChatGPT  & \textbf{57.49} & \textbf{88.68}  & 23.23 \\
                \midrule
                Ours & 33.00 & 35.89 & \textbf{28.90} \\
    		\bottomrule
		\end{tabular}
        \caption{The results on the subsets of the dev sets from the out-of-domain ODQA datasets.}
	\label{tab:out-domain}
\end{table}

\subsection{Main Results}
\paragraph{Results on In-domain Datasets.}
Table~\ref{tab:fb-res} and Table~\ref{tab:wiki-res} show the results on in-domain datasets based on Freebase and Wikidata, respectively. 
First, LM-based seq2seq generation methods can achieve better F1 score compared to the subgraph-based reasoning methods on the WebQSP and KQA Pro. It indicates that the SPARQL query generated by the LM can obtain a more complete answer set, and the structured query can better support some complex operations~(\eg maximum, count) than the traditional subgraph-based reasoning methods. 
Second, although LLMs are powerful, directly using Davinci-003, ChatGPT, and even GPT-4 still has a large performance gap compared with the best fine-tuned methods in WebQSP, GrailQA, and KQA Pro, indicating the difficulty of answering complex questions solely by LLMs. 

Finally, our KG-Agent is substantially better than all other competitive baselines in all datasets after instructing tuning on the mixed data. With the mutual augmentation between different datasets, our approach achieves 1.7\%, 7.5\%, and 2.7\% improvements of F1 on WebQSP, CWQ, and Grailqa, respectively. 
Benefiting from the autonomous reasoning mechnism, our approach can perform reasoning on the two KGs and obtain consistent improvement on all datasets.

\paragraph{Results on Out-of-domain Datasets.}
After instruction tuning, we directly evaluate the zero-shot performance of our KG-Agent on the out-of-domain datasets. As shown in Table~\ref{tab:out-domain}, although fine-tuned with full data, the small pre-trained language models~(\eg T5 and BART) can not effectively answer these factual questions. Owing to the large-scale parameters, Davinci-003 and ChatGPT performs well on NQ and TQ, which are constructed based on Wikipedia, the corpus that they may have been pre-trained on.
However, they perform not well on WQ, which is constructed based on Freebase KG.
In contrast, our KG-Agent only needs to learn how to interact with KG instead of memorizing the specific knowledge. Thus, it can utilize the external KG in zero-shot setting, and achieve consistent improvement compared to fine-tuned pre-trained language models.
\begin{table}[t]
    \renewcommand\arraystretch{1}
    \setlength\tabcolsep{4.3pt}
	\centering
	\small
            \begin{tabular}{l  c  c c }
			\toprule
			\textbf{Models} & \textbf{MQA-1hop} & \textbf{MQA-2hop} & \textbf{MQA-3hop} \\
			\midrule
			GraftNet & 82.5    & -  &  -  \\
			EmbedKGQA & 92.0    & 40.7  &  34.6  \\
			  NSM & 94.8  &   97.0    & 91.0 \\
                TransferNet & 96.5    & 97.5  &  90.1 \\
                \midrule
			ChatGPT & 61.9 & 31.0 & 43.2\\
			StructGPT  & 94.2 & 93.9 & 80.2  \\
                \midrule
                Ours & \textbf{97.1} & \textbf{98.0} & \textbf{92.1} \\
    		\bottomrule
		\end{tabular}
        \caption{The results on the three subsets of MetaQA. We copy the results of baselines from~\citet{Jiang-2023-arxiv-StructGPT}.}
	\label{tab:res-metaqa}
\end{table}
\subsection{Further Analysis}

\paragraph{Transfer to Domain-specific KG.}
To evaluate the transferability of our approach on other KGs, we test our KG-Agent on the MetaQA dataset which is based on a movie domain KG. 
Following existing work~\cite{NSM, Jiang-2023-arxiv-StructGPT}, we show the one-shot results on the test set in Table~\ref{tab:res-metaqa}. ChatGPT performs not well when directly answering these domain-specific questions, where the performance drops 45\% absolutely on the MQA-3hop subset compared to the supervised fine-tuned TransferNet model. After equipping the LLM with the KG, StructGPT can greatly outperform ChatGPT with about 37\% improvement. In contrast, our KG-Agent can obtain consistent performance improvement compared to the competitive supervised fine-tuning baselines on all subsets. It indicates that the agent indeed learns the general ability about reasoning on KG, which can be efficiently transferred to other KGs.

\begin{figure}[t]
	\centering
	\subfigure{
		\centering
		\includegraphics[width=0.22\textwidth]{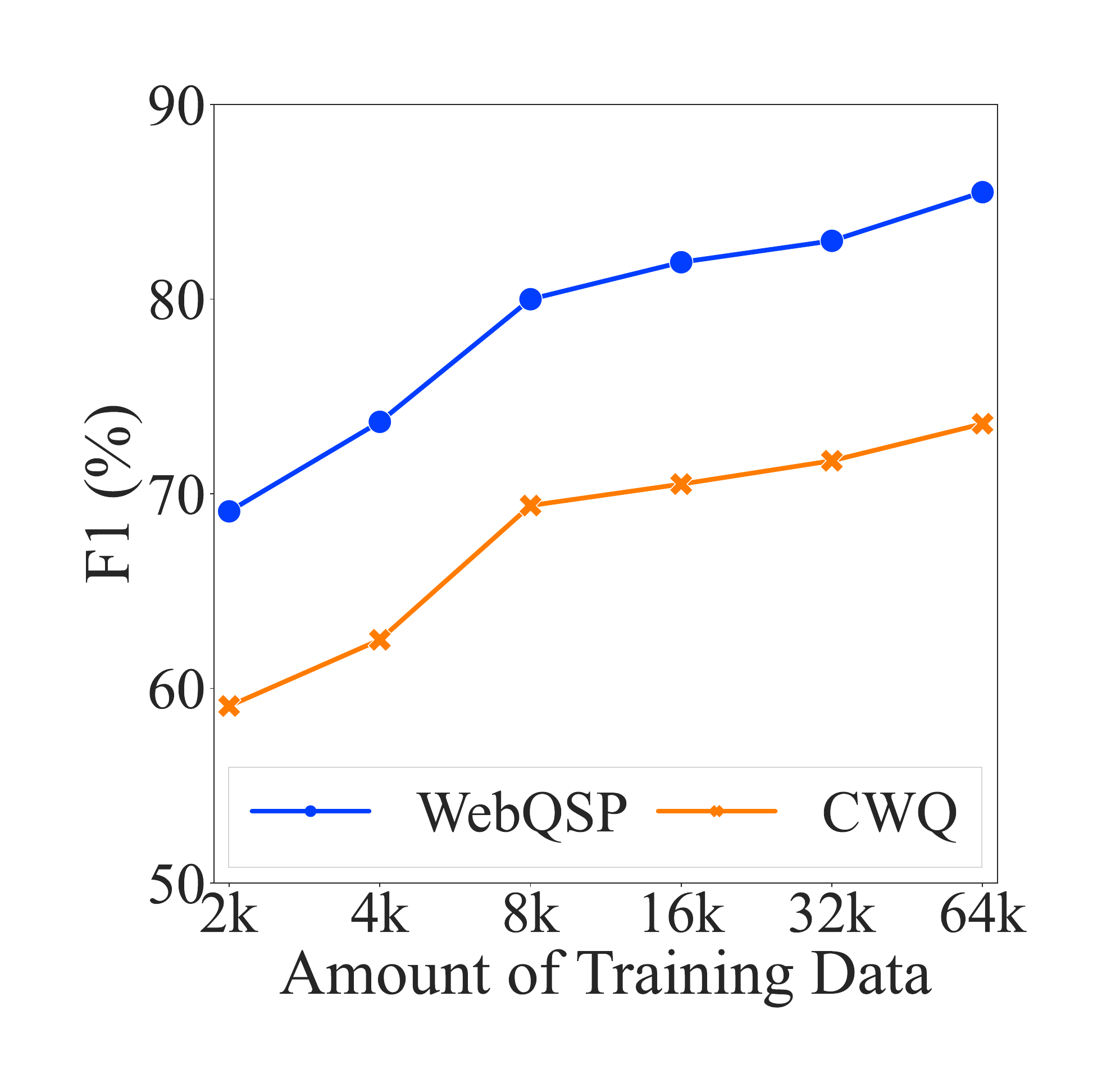}
	}
	\subfigure{
		\centering
		\includegraphics[width=0.22\textwidth]{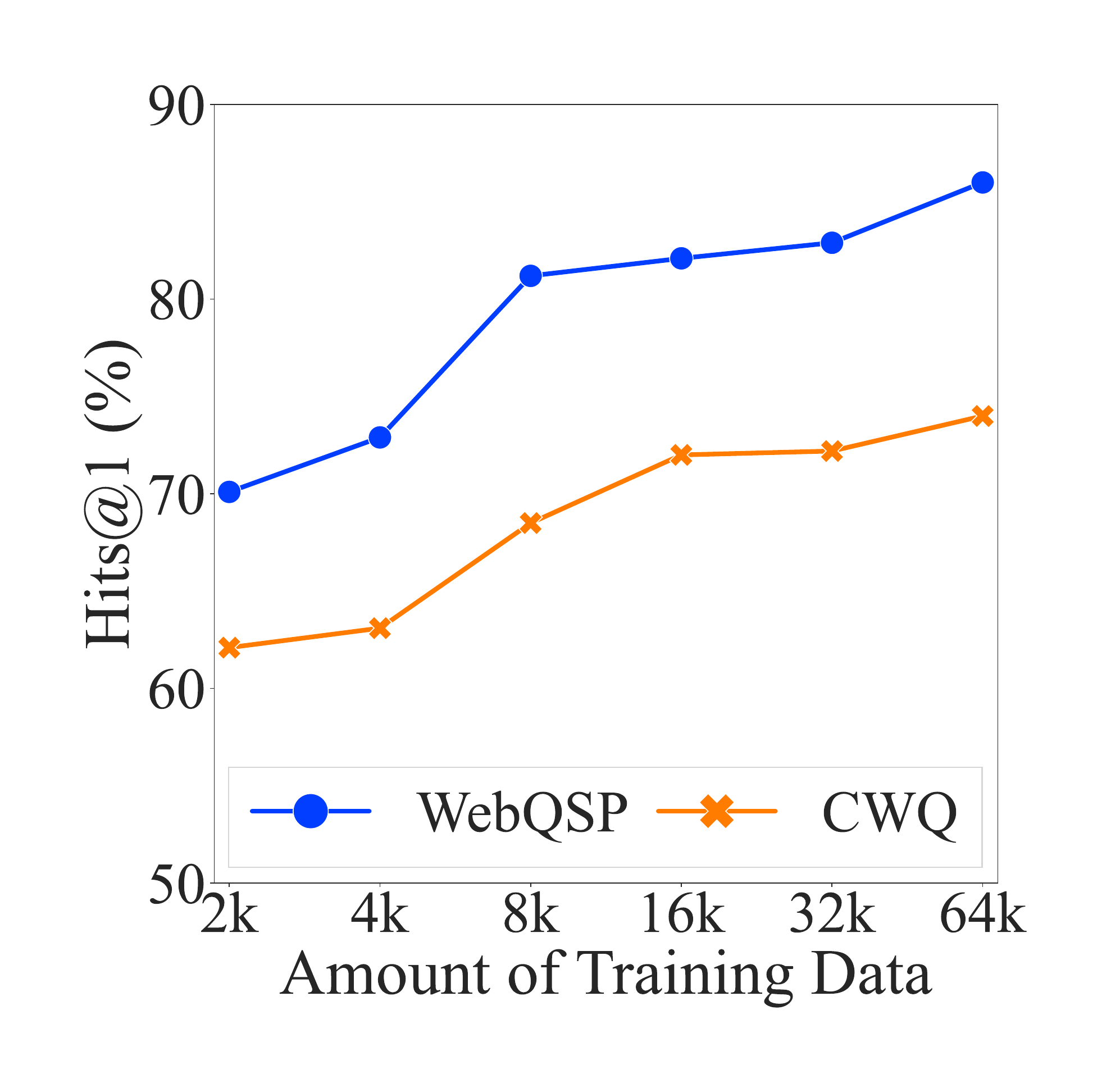}
	}
	\centering
	\caption{The F1~(Left) and Hits@1~(Right) scores of KG-Agent on the test set of WebQSP and CWQ with a various amount of instruction tuning data.}
	\label{fig:amount}
\end{figure}

\paragraph{Effect of Instruction Amount.}
We explore how the amount of instructions affects the performance of KG-Agent and show the results in Figure~\ref{fig:amount}. With a constant sampling proportion, we scale the total amount from 2k to 64k in an exponential way and evaluate the F1 and Hist@1 scores on WebQSP and CWQ datasets. As we can see, the performance increases with more instruction tuning data, and eventually reaches a stable state, which indicates the importance of data amount. At the same time, with the data amount increasing from 16k to 64k, the KG-Agent doesn't obtain a remarkable performance improvement. We think this is relevant to the variety of our instruction tuning data, which is illustrated in existing work~\cite{Chung-arxiv-2022-Scaling, Aribandi-ICLR-2022-ExT5}. Therefore, we will construct more diverse samples in the future to further boost the performance.

\begin{table}[t]
	\centering
	\small
    \begin{tabular}{c | c c c | c}
        \toprule
          Proportion& WebQSP & CWQ & GrailQA & Average\\
        \midrule
            1:10:5 & 80.0& 69.8& 86.1 & \textbf{78.6}  \\
            \underline{2}:10:5 & 81.2& 68.7& 83.3 & 77.8 \\
            1:\underline{20}:5 & 78.9.& 73.6& 78.8 & 77.1 \\
            1:10:\underline{10} & 80.8& 66.9& 84.3 & 77.3 \\
        \bottomrule
    \end{tabular}
        \caption{The F1 scores on three in-domain datasets after instruction tuning under different sampling proportions. We highlight the changed proportion with an underline. }
        \label{tab:ablation}
\end{table}

\paragraph{Effect of Tuning Data Proportion.}
Our experiment finds that only sampling 10K samples from existing datasets is enough for backbone LLM to learn the autonomous decision making capability.
Here, we conduct a further ablation study to explore the impact of sampling proportion on the agent's performance when keeping the total amount of instruction tuning data constant.
Specifically, we evaluate the agent performance of WebQSP, CWQ, and GrailQA when doubling the proportion of one dataset while maintaining the other two dataset proportions. We show the results in Table~\ref{tab:ablation}. We can see that as the sampling proportion of a certain dataset increases, the agent performance on it consistently improves. However, for the average performance on all three datasets, all variants are lower than our selected proportion, indicating that the proportion we chose is suitable for the LLM to balance and master more comprehensive and general abilities.

\section{Conclusion}
\label{conclusion}

In this work, we proposed an autonomous agent framework to synergize LLMs and KGs to perform complex reasoning over KG, namely KG-Agent. In our approach, we first curated a toolbox for KG, consisting of three types of tools to support the typical operations when reasoning on KG. Then, we developed an autonomous iteration mechanism based on tool selection and memory updation that integrates the LLM, multifunctional toolbox, KG-based executor, and knowledge memory, for reasoning over KG. Next, we leveraged existing KGQA datasets to synthesize the code-based instruction tuning dataset. Finally, with only 10K tuning samples, we implemented the autonomous agent relying on the smaller 7B LLM, which mostly outperforms state-of-the-art baselines based on full-data tuning or larger LLMs. In future work, we will consider extending our framework to deal with more types of structured data, \eg databases and tables.

\section*{Limitations} \label{limitations}
Although KG-Agent demonstrates remarkable performance across various complex factual question answering tasks, there are some limitations of our method.
First, we only use the LLaMA2-7B as the backbone LLM, which has a strong capability after instruction tuning. Hence, more experiments are required to evaluate other LLMs with comparable parameter sizes, such as Mistral-7B~\cite{Mistral} or CodeLLaMA-7b~\cite{CodeLLaMA}.
Second, we focus on reasoning over the KG to answer the factual questions. We should consider extending our framework to deal with more types of knowledge sources, \eg databases or tables.
Third, we only evaluate factual question answering tasks based on KG. Future work should include wider evaluation scenarios to evaluate the universality of our method, \eg data-to-text and formal-language-to-text~\cite{xie-EMNLP-22}. 
Finally, we have tried our best to tune the LLM only to answer the questions based on the KG information, and avoid generating discriminatory and risky responses for user questions. However, we should add more rule-based methods to post-process the predictions and filter the illegal responses.

\bibliography{custom}
\bibliographystyle{acl_natbib}

\newpage
\clearpage
\appendix
\label{appendix}

\section{Experiment Setup}
\label{sec:app-experiment}
\subsection{Datasets} 
We select four popular complex KGQA datasets as in-domain datasets, \ie \textit{WebQuestionsSP (WebQSP)}~\cite{WebQSP}, \textit{Complex WebQuestions 1.1 (CWQ)}~\citep{CWQ}, and \textit{GrailQA}~\citep{GrailQA}, which are based on Freebase, and \textit{KQA Pro}~\cite{KQA-Pro}, which is based on Wikidata.
And we select three representative ODQA datasets as out-domain datasets, which are \textit{WebQuestions~(WQ)}~\cite{Berant-WQ-ACL-13}, \textit{Natural Questions~(NQ)}~\cite{Chen-NQ-ACL-17}, and \textit{TriviaQA~(TQ)}~\cite{Joshi-TQ-ACL-17}. Since we only rely on the KG to answer questions, we filter the questions in ODQA datasets that can not be linked to any entity in KG, denoted as \textit{WQ-Freebase}, \textit{NQ-Wiki}, and \textit{TQ-Wiki}, respectively.
Besides, we further select the \textit{MetaQA}~\cite{zhang-aaai-18}, which is based on a domain-specific movie KG, to evaluate the generalibility of our method.
The detail description of these selected datasets is as follows:

$\bullet$ \textbf{WebQSP} consists of 4,737 questions. The answer entities are within a maximum of 2 hops from the topic entity on the Freebase KG. We adopt the train/valid/test splits from GraftNet~\citep{GraftNet} for consistency.

$\bullet$ \textbf{CWQ} is constructed based on WebQSP, which is more challenging. It complicates WebQSP by extending the question entities or adding constraints to restrict the answers. The answer entities are within a maximum of 4 hops from the topic entity on the Freebase KG.

$\bullet$ \textbf{GrailQA} consists of 64,331 questions. Compared to WebQSP and CWQ, it focuses on a more comprehensive generalization capability evaluation from three levels~(\ie i.i.d, compositional, and zero-shot).

$\bullet$ \textbf{KQA Pro} consists of 117,970 questions. The above three datasets are based on Freebase, and it is based on Wikidata, and require multiple reasoning capabilities including compositional reasoning, multi-hop reasoning, quantitative comparison, set operations, and etc.

$\bullet$ \textbf{MetaQA} comprises over 400,000 questions based on a movie domain KG, with answer entities located up to three hops away from
the topic entities. Based on the number of hops, the dataset is divided into three sub-datasets: MetaQA-1hop, MetaQA-2hop, and MetaQA-3hop. Following existing work~\cite{NSM}, we randomly sample just one training case for each question template from the original training set, to form a one-shot training dataset.

$\bullet$ \textbf{WQ} consists of 6,642 questions. The questions are mostly centered around a single named entity and are supposed to be answerable by Freebase KG. We extract xx questions from the original test set to compose the WQ-freebase subset.

$\bullet$ \textbf{NQ} consists of 323,045 questions. Each example contains a question from the Google search and the corresponding answers, which are text spans on the Wikipedia page. Following existing work~\cite{pack-knowledge}, we use the open version of this dataset which discards answers with more than 5 tokens. We extract xx questions from the original test set to compose the NQ-Wiki subset.

$\bullet$ \textbf{TQ} consists of 110K questions. Each example contains a question authored by trivia enthusiasts, and the answers are text spans from the Web or Wikipedia. Following existing work~\cite{pack-knowledge}, we use its unfiltered version for evaluation. We extract xx questions from the original test set to compose the TQ-Wiki subset.

\subsection{Evaluation Protocol} For KGQA, following existing work~\cite{GraftNet}, we use Hits@1 and F1 metrics for WebQSP and CWQ datasets, F1 metric for GrailQA dataset, and Hits@1 for MetaQA. The Hits@1 evaluates the correctness of the top-ranked answer while F1 considers coverage of all the predicted answers.
It's worth noting that some baselines and our approach would return all the unordered answers at the end, which is not suitable for the Hist@1 metric. For a comprehensive comparison, we randomly select one answer per question as the top-ranked answer and then calculate the average Hits@1 result by repeating this process 100 times following existing work~\cite{TIARA}.
For ODQA, following existing work~\cite{pack-knowledge}, we report the EM metric, which evaluates whether the predicted answer is the same as the gold one after performing normalization.

\subsection{Baselines for Comparison} 
For KGQA, we consider the following three types of baseline methods for performance comparison:

$\bullet$ \textbf{subgraph-based reasoning} methods which perform answer reasoning in a retrieval subgraph form KG, including GrafeNet~\cite{GraftNet}, NSM~\cite{NSM}, SubgraphRetrieval~\cite{Zhang-ACL-22}, UniKGQA~\cite{jiang-arxiv-22}, and ReasoningLM~\cite{ReasoningLM} for datasets on Freebase, and KVMemNet~\cite{miller-EMNLP-16}, EmbedKGQA~\cite{Saxena-ACL-20}, and RGCN~\cite{RGCN} for datasets on Wikidata;

$\bullet$ \textbf{LM-based seq2seq generation} methods which generate the final SPARQL query by fine-tuning a sequence-to-sequence language model, including RNG-KBQA~\cite{RNG-KBQA}, Uni-Parser~\cite{UniParser}, ArcaneQA~\cite{ArcaneQA}, PanGu w/ T5-3B~\cite{Gu-ACL-23}, TIARA~\cite{TIARA}, and FC-KBQA~\cite{FC-KBQA} for datasets on Freebase, and RNN SPARQL and BART SPARQL~\cite{KQA-Pro} for datasets on Wikidata;

$\bullet$ \textbf{LLM-based} methods which utilize the powerful zero-shot or few-shot capabilities of LLMs to answer the question without fine-tuning, including ROG~\cite{Luo-arxiv-23-Reasoning}, StructGPT~\cite{Jiang-2023-arxiv-StructGPT}, gpt-3.5-turbo-instruct~(Davinvi-003)~\footnote{https://platform.openai.com/docs}, gpt-3.5-turbo~(ChatGPT)~\footnote{https://platform.openai.com/docs}, and gpt-4~(GPT-4)~\footnote{https://platform.openai.com/docs} for both in-domain datasets.

For ODQA, we focus on the closed-book setting where no documents are provided and consider the following two types of baseline methods:

$\bullet$ \textbf{Fine-tune based} methods which learn to predict the answers, including T5-Base, T5-Large, BART-base, and BART-Large from ~\citep{pack-knowledge};

$\bullet$ \textbf{LLM-based} methods which directly answer the questions in zero-shot setting, including gpt-3.5-turbo-instruct~(Davinvi-003) and gpt-3.5-turbo~(ChatGPT).

\subsection{Implementation Details}
For instruction tuning data construction, we randomly sample a total of 10,000 training data from in-domain datasets in a ratio of 1:5:5:10 for WebQSP, KQA Pro, GrailQA, and CWQ according to some prior empirical studies.
Since we focus on the reasoning process over KG, we suppose the entities have been given for each question following existing work~\cite{GraftNet, NSM, Jiang-2023-arxiv-StructGPT}.
For instruction tuning, we use the LLaMA2-7B~\cite{LLaMA2} as our backbone LLM. We use a cosine learning rate schedule with an initial learning rate of \text{2e-5}, a weight decay of 0.1, a batch size of 256, a maximum length of 1500, and finally fine-tune the model for 3 epochs. 
For the relation retrieval model and entity disambiguation model in the semantic tool, we build them following the existing work~\cite{Zhang-ACL-22, TIARA}.

After instruction tuning, for in-domain datasets, we evaluate the performance of our KG-Agent on the test set of CWQ, WebQSP, KQA Pro, and the dev set of GrailQA.
For out-domain datasets, we evaluate the zero-shot performance of our KG-Agent on the NQ-Wiki, TQ-Wiki, and WQ-Freebase.
For the domain specific dataset, \ie MetaQA, we follow existing work~\cite{NSM, Jiang-2023-arxiv-StructGPT} to extract the one-shot tuning subset from the original training set and fine-tune our KG-Agent with it.
When evaluating the performance of Davinci-003, ChatGPT, and GPT4, we use the latest February version of APIs from OpenAI. And for in-domain datasets, we provide six demonstrations for each test question and parse the prediction results following existing work~\cite{ToG,Jiang-2023-arxiv-StructGPT}, we show the prompt with demonstration for each dataset in Table~\ref{tab:icl_example}.
For the selection of demonstrations, we randomly sample from the corresponding training set for each dataset.
For out-domain datasets, since they are open-domain question answering tasks, we directly input the question to LLMs with proper prompt, as shown in Table~\ref{tab:icl_example}.

\begin{table*}[t]
    \small
    \centering
    \begin{tabular}{c | l}
    \toprule
    \textbf{Dataset} & \textbf{Prompt} \\
    \midrule
    \multirow{20}{*}{\textbf{WebQSP}}  & Question: where is the syracuse university? \\
    & Answer: [New York | Syracuse | United States of America].\\
    & Question: where is the mtv headquarters?\\
    & Answer: [New York City].\\
    & Question: what are the 3 official languages of spain?\\
    & Answer: [Spanish Language].\\
    & Question: what timezone is new england usa in?\\
    & Answer: [Eastern Time Zone].\\
    & Question: who started southwest airlines?\\
    & Answer: [Herb Kelleher | Rollin King].\\
    & Question: what was irving langmuir famous for?\\
    & Answer: [Scientist].\\
    & Question: \{test question\} \\
    & Answer:\\
    \midrule
    \multirow{20}{*}{\textbf{CWQ}}  & Question: Who is the president in the place where the government of Peru is located? \\
    & Answer: [Ollanta Humala].\\
    & Question: Where did Martin Luther King attend university, that has less than 2,586 undergraduates?\\
    & Answer: [Morehouse College].\\
    & Question: What movie produced by the company New Line Cinema was Taylor Lautner in?\\
    & Answer: [Valentine's Day].\\
    & Question: Which year did the team that plays at Turner Field win the World Series?\\
    & Answer: [1995 World Series].\\
    & Question: Which airports are in the circulation area of Il Manifesto?\\
    & Answer: [Leonardo da Vinci–Fiumicino Airport | Ciampino–G. B. Pastine International Airport].\\
    & Question: What were the professions held by the publisher of "The Awakening?"?\\
    & Answer: [Businessperson | Novelist | Writer | Author].\\
    & Question: \{test question\} \\
    & Answer:\\
    \midrule
    \multirow{20}{*}{\textbf{GrailQA}}  & Question: what does the thiokol rocket do? \\
    & Answer: [Launch vehicle].\\
    & Question: what is the club interest of inverness yacht club?\\
    & Answer: [Sailing].\\
    & Question: who is the tour operator of kiribati?\\
    & Answer: [Fly Water Adventures | Kiribati Holidays | Otintaai Tours | Molloy's Tours].\\
    & Question: 1998 marsala vergine terre arse contains what type of grapes?\\
    & Answer: [Catarratto | Grillo | Ansonica].\\
    & Question: \makecell[l]{how many ice hockey coaches have coached the team \\that is currently coached by the eisbaren berlin?}\\
    & Answer: [1].\\
    & Question: court of appeal of sri lanka has what inferior court?\\
    & Answer: [Supreme Court of Sri Lanka].\\
    & Question: \{test question\} \\
    & Answer:\\
    \midrule
    \multirow{20}{*}{\textbf{KQA Pro}}  & Question: Which website officially represents Morgan Creek Productions? \\
    & Answer: [http://www.morgancreek.com/].\\
    & Question: \makecell[l]{Which is shorter: The Killers, with a story set in Los Angeles, \\or Sherlock Holmes, produced by 20th Century Fox?}\\
    & Answer: [Sherlock Holmes].\\
    & Question: What is the street address for the University of San Diego?\\
    & Answer: [5998 Alcala Park, San Diego, CA, 92110-2492].\\
    & Question: How is the Francis Bacon who died in New Haven related to the Yale School of Medicine?\\
    & Answer: [educated at].\\
    & Question: For the film titled Aladdin, where is it published on its publication date of 2019-05-24?\\
    & Answer: [United States of America].\\
    & Question: Who wrote The Postman which was published in 1985?\\
    & Answer: [David Brin].\\
    & Question: \{test question\} \\
    & Answer:\\
    \midrule
    \textbf{\makecell{NQ-Wiki \\ TQ-Wiki \\WQ-Freebase}} & Answer the following question with one or few words. Question: \{test question\}\\
    \bottomrule
    \end{tabular}
    \caption{The prompts used for each dataset when evaluating the ChatGPT, Davinci-003, and GPT-4 models. When performing evaluation, just replace the ``\{test question\}'' with the test question.}
    \label{tab:icl_example}
\end{table*}

\begin{table*}[t]
    \small
    \centering
    \begin{tabular}{c l c}
    \toprule
    \textbf{Type} & \textbf{Tool}  & \textbf{Description}\\
    \midrule
    \multirow{18}{*}{\makecell{Extraction \\ Tool}} & \multirow{2}{*}{get$\_$relation} & Input: entity set $\{e\}$ $\rightarrow$ Output: one-hop relations $R_{\{e\}}$ \\
    & & Return the incoming and outgoing relations of the given entity set $\{e\}$ on KG. \\
    \cmidrule[0.001pt]{2-3}
    & \multirow{2}{*}{get$\_$head$\_$entity} & Input: entity set $\{e\}$, relation $r$ $\rightarrow$ Output: entity set $\{e\}$ \\
    & & Return the head entity set of the given tail entity set $\{e\}$ along the relation $r$. \\
    \cmidrule{2-3}
    & \multirow{2}{*}{get$\_$tail$\_$entity} & Input: entity set $\{e\}$, relation $r$ $\rightarrow$ Output: entity set $\{e\}$ \\
    & & Return the tail entity set of the given head entity set $\{e\}$ along the relation $r$. \\
    \cmidrule{2-3}
    & \multirow{2}{*}{get$\_$entity$\_$by$\_$type} & Input: string type $t$ $\rightarrow$ Output: entity set $\{e\}$ \\
    & & Return the entity set belonging to the given type $t$. \\
    \cmidrule{2-3}
    & \multirow{6}{*}{get$\_$entity$\_$by$\_$constraint} & Input: entity set $\{e\}$, relation $r$, operator $o$, string value $v$ $\rightarrow$ Output: entity set $\{e\}$ \\
    & & \makecell[c]{Return the new entity set whose tail entity along $r$ satisfies the constraint condition. \\If $v$ is not empty, the $o$ should be one of \{``='',``>'',``>='',``<'',``<=''\}, which means \\the comparison between the tail entity and string value should satisfy the operator. \\Else, the $o$ should be one of \{``argmax'',``argmin''\}, which means the tail entity \\should be the maximum or minimum value.}\\
    \cmidrule{2-3}
    & \multirow{2}{*}{get$\_$candidate$\_$entity} & Input: string entity mention $m$ $\rightarrow$ Output: entity set $\{e\}$ \\
    & & Return the candidate linked entity set on the KG for the given entity mention $m$. \\
    \midrule
    \multirow{13}{*}{\makecell{Logic \\ Tool}} & \multirow{2}{*}{count} & Input: entity set $\{e\}$ $\rightarrow$ Output: integer \\
    & & Return the number of entities in the given entity set $\{e\}$. \\
    \cmidrule{2-3}
    & \multirow{2}{*}{intersect} & Input: entity set list $[\{e\}]$ $\rightarrow$ Output: entity set $\{e\}$ \\
    & & Return the intersection of the given list of entity sets.  \\
    \cmidrule{2-3}
    & \multirow{2}{*}{union} & Input: entity set list $[\{e\}]$ $\rightarrow$ Output: entity set $\{e\}$ \\
    & & Return the union of the given list of entity sets.  \\
    \cmidrule{2-3}
    & \multirow{2}{*}{judge} & Input: entity set $\{e\}$, relation $r$, operator $o$, string value $v$ $\rightarrow$ Output: boolean \\
    & & \makecell[c]{Return a boolean value indicating whether the comparison between the tail entity of \\the given entity set $\{e\}$ along relation $r$ and the given value $v$ satisfies the operator $o$.} \\
    \cmidrule{2-3}
    & \multirow{2}{*}{end} & Input: entity set $\{e\}$ $\rightarrow$ Output: entity set $\{e\}$ \\
    & & Return the entity set as the final answer and end the reasoning process. \\
    \midrule
    \multirow{7}{*}{\makecell{Semantic \\ Tool}} & \multirow{3}{*}{retrieve\_relation} & Input: relation set $\{r\}$ $\rightarrow$ Output: relation set $\{r\}$ \\
    & & \makecell[c]{Retrieve relations from the given relation set $\{r\}$ that are \\semantically relevant to the question through neural network.}  \\
    \cmidrule{2-3}
    & \multirow{2}{*}{disambiguate\_entity} & Input: entity set $\{e\}$ $\rightarrow$ Output: entity $e$ \\
    & & \makecell[c]{Disambiguate the candidate linked entity $\{e\}$ based on the question semantics \\and entity information on KG~(\eg one-hop relations) through neural network.} \\
    \bottomrule
    \end{tabular}
    \caption{The detailed definition and usage of all the tools.}
    \label{tab:toolbox}
\end{table*}

\section{Summary of Toolbox}
\label{sec:app-tool}
We summarize the tool name, tool description, and the input argument and output of tools in Table~\ref{tab:toolbox}.

\clearpage

\end{document}